\documentclass[]{cit_lab_mfr}

\usepackage{hyperref}
\usepackage{cleveref}
\usepackage{verbatim}

\usepackage{wrapfig}  
\usepackage{graphicx}
\usepackage{floatrow}
\usepackage{subcaption}
\usepackage{listings}
\usepackage{algorithm}

\usepackage{lineno,hyperref}
\usepackage{multirow}
\usepackage{amsmath,amsfonts}
\usepackage{booktabs}
\usepackage{algorithm}
\usepackage{algorithm}
\usepackage{algorithmic}

\usepackage{subcaption} %

\usepackage[toc,page,header]{appendix}

\usepackage{minitoc}

\title{\textsc{MedAide}: Information Fusion and Anatomy of Medical Intents via LLM-based Agent Collaboration}

\author{%
\parbox{\textwidth}{\centering
Dingkang Yang$^{1,2,*,\dagger}$, Jinjie Wei$^{,1,2,3,*}$, Mingcheng Li$^{1,2,*}$, Jiyao Liu$^{3,4}$, Lihao Liu$^{3}$, Ming Hu$^{3}$, Junjun He$^{3}$, Yakun Ju$^{5}$, Wei Zhou$^{6}$, Yang Liu$^{7,\S}$, Lihua Zhang$^{1,8,9,10,\S}$
}}

\affiliation{%
\parbox{\textwidth}{\centering\small
$^1$College of Intelligent Robotics and Advanced Manufacturing, Fudan University$\,$
$^2$Multimodal Foundation Research Team, CIT Lab\\[1mm]
$^3$Shanghai Artificial Intelligence Laboratory$\,$
$^4$Institute of Science and Technology for Brain-Inspired Intelligence, Fudan University\\[1mm]
$^5$School of Computing and Mathematical Sciences, University of Leicester$\,$
$^6$Cardiff University\\[1mm]
$^7$Department of Computer Science, The University of Toronto$\,$
$^8$Institute of Metaverse \& Intelligent Medicine\\[1mm]
$^9$Jilin Provincial Key Laboratory of Intelligence Science and Engineering\\[1mm]
$^{10}$Engineering Research Center of AI and Robotics, Ministry of Education\\[1mm]
}}

\contribution[*]{Equal contribution}
\contribution[\dagger]{Project lead}
\contribution[\S]{Corresponding Author}

\renewcommand{\thefootnote}{\fnsymbol{footnote}}
\renewcommand{\thefootnote}{\arabic{footnote}}  

\abstract{
In healthcare intelligence, the ability to fuse heterogeneous, multi-intent information from diverse clinical sources is fundamental to building reliable decision-making systems. Large Language Model (LLM)-driven information interaction systems currently showing potential promise in the healthcare domain. Nevertheless, they often suffer from information redundancy and coupling when dealing with complex medical intents, leading to severe hallucinations and performance bottlenecks.
To this end, we propose \textsc{MedAide}, an LLM-based medical multi-agent collaboration framework designed to enable intent-aware information fusion and coordinated reasoning across specialized healthcare domains.
Specifically, we introduce a regularization-guided module that combines syntactic constraints with retrieval-augmented generation to decompose complex queries into structured representations, facilitating fine-grained clinical information fusion and intent resolution.
Additionally, a dynamic intent prototype matching module is proposed to utilize dynamic prototype representation with a semantic similarity matching mechanism to achieve adaptive recognition and updating of the agent's intent in multi-round healthcare dialogues.
Ultimately, we design a rotation agent collaboration mechanism that introduces dynamic role rotation and decision-level information fusion across specialized medical agents.
Extensive experiments are conducted on four medical benchmarks with composite intents. Experimental results from automated metrics and expert doctor evaluations show that \textsc{MedAide} outperforms current LLMs and improves their medical proficiency and strategic reasoning.
}
\date{\today}
\checkdata[Corresponding]{\url{dicken@fysics.ai}}
\checkdata[Project Page]{\url{https://github.com/ydk122024/MedAIDE}}

\begin{document}
\maketitle

\begingroup 
\renewcommand{\thefootnote}{\fnsymbol{footnote}}
\endgroup

\section{Introduction}
The emergence of information fusion technologies has revolutionized healthcare systems by enabling the integration of multi-source medical data for enhanced decision-making capabilities~\cite{shaik2024survey}. Contemporary multi-source information fusion approaches have proven crucial for processing diverse healthcare data modalities, enabling comprehensive patient assessment and personalized treatment strategies~\cite{li2024deep}. In the healthcare domain, the integration of artificial intelligence with information fusion techniques has demonstrated significant potential for smart healthcare applications~\cite{ali2020smart}. The advancement of multimodal medical signal fusion has played a crucial role in smart healthcare systems, enabling real-time analysis and providing medical professionals with enhanced insights for disease detection and prediction accuracy~\cite{muhammad2022comprehensive}. Furthermore, multi-modal lifelog data fusion approaches have shown promising results in improving human activity recognition and health monitoring applications~\cite{gao2024multi}. Advanced deep learning technologies \cite{ChatGPT,o1-preview, wang2025vgr, lei2025scalability, yang2024pediatricsgpt, wei2025learning, liu2024generalized, liu2023amp} have shown exceptional generalization capabilities in general scenarios for multiple purposes within human-machine interactions.
Developing goal-oriented conversation systems~\cite{cheng2024cooper} has received increasing attention in recent years, particularly in medical contexts where information fusion enables sophisticated reasoning across heterogeneous data sources. In this context, LLM-centered interactive medical assistants~\cite{yang2024pediatricsgpt,yang2024zhongjing,chen2023meditron70b} have become research hotspots that promise to improve diagnosis efficiency and promote service automation. The advancement of multimodal medical signal fusion has played a crucial role in smart healthcare systems, enabling real-time analysis and providing medical professionals with enhanced insights for disease detection and prediction accuracy~\cite{muhammad2022comprehensive}.
Previous attempts have infused LLMs with healthcare-specific knowledge through relevant corpus construction and multi-stage training procedures . However, traditional approaches often overlook the potential of multi-source information fusion for achieving a comprehensive understanding of complex medical scenarios~\cite{gao2024multi}. Although these strategies enhance the models' comprehension of medically relevant intents, bottlenecks remain when faced with real-world applications that require sophisticated reasoning and accurate feedback, particularly in scenarios that demand the integration of diverse information sources for optimal clinical decision-making.

Given the mimicking of learned feedback in human behaviors, automated medical agent construction promises to enhance LLMs' instruction following and logical profiling capabilities.
The collaboration among multiple agents to handle different patient inquiries and symptomatic caseloads facilitates accurate dialog goal fulfillment while accounting for individual differences~\cite{fan2024aihospitalbenchmarkinglarge}.
Despite significant advancements, current efforts focus primarily on medical education training~\cite{wei2024medco} or selective question-answering~\cite{tang2024medagentslargelanguagemodels}, 
lacking a comprehensive understanding when dealing with the sophisticated intents behind user queries in real-world diagnosis and treatment scenarios. In addition, the agents in most collaborative frameworks are usually familiar with only limited intents, making it difficult to provide systematic recommendations.

\begin{figure}[t]
  \centering
  \includegraphics[width=0.8\linewidth]{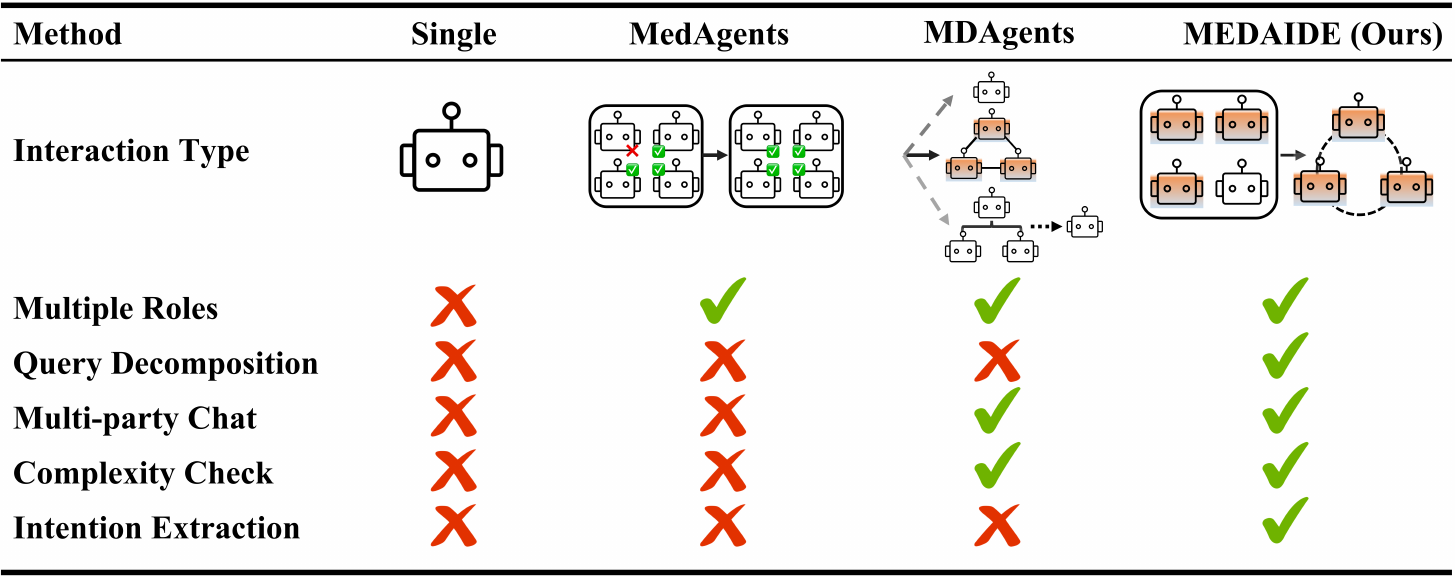}
  \caption{Comparison between our framework and previous methods. Among these
works, \textsc{MedAide} is the only one with intent awareness and comprehensive analysis capabilities.
  }
  \label{demo}
\end{figure}

To address these issues, we propose \textsc{MedAide}, an LLM-based medical multi-agent collaboration framework designed to enable intent-aware information fusion and coordinated reasoning across specialized healthcare domains.
There are three core contributions in \textsc{MedAide} based on the tailored components. 
Specifically, we propose a regularization-guided information extraction module, which fuses syntactic constraints with retrieval-augmented generation to transform compositional medical queries into structured semantic representations. This structured decomposition enhances the granularity and reliability of clinical knowledge extraction, facilitating downstream intent interpretation. In addition, a dynamic intent prototype matching mechanism is introduced to adaptively recognize and update agent intent through semantic similarity-based alignment during multi-turn healthcare dialogues. 
Ultimately, we design a rotation-based agent collaboration mechanism that enables dynamic role-switching and decision-level information fusion across agents, further ensuring coherent and globally optimized decision-making.
Experimental results on four medical benchmarks with composite intents, evaluated by both automated metrics and expert physicians, demonstrate that \textsc{MedAide} significantly improves medical reasoning accuracy, intent alignment, and agent coordination over strong LLM baselines.

\begin{figure*}[t]
  \centering
  \includegraphics[width=1.0\linewidth]{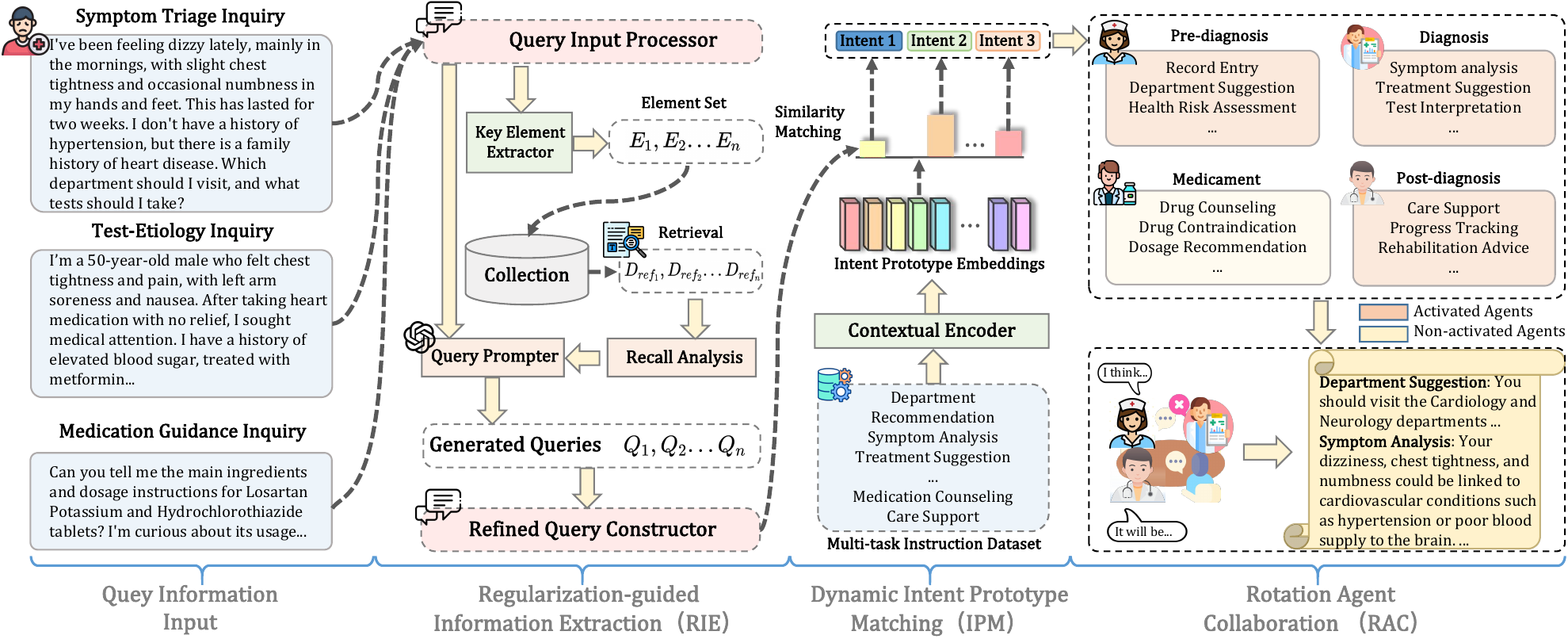}
  \caption{Illustration of the \textsc{MedAide} framework.
  \textsc{MedAide} consists of three major phases: (i) Regularization-guided Information Extraction (RIE) employs syntactic parsing and retrieval-augmented generation to reformulate complex medical queries and extract clinical elements. (ii) Dynamic Intent Prototype Matching (IPM) utilizes a BioBERT-based encoder to generate intent embeddings for precise classification across 17 medical intent categories. (iii) Rotation Agent Collaboration (RAC) implements a polling-based mechanism where specialized agents (pre-diagnosis, diagnosis, medicament, post-diagnosis) dynamically rotate as main contacts, coordinating multi-source information fusion through cross-agent knowledge integration.
  }
  \label{arc}
\end{figure*}

\section{Related Work}
\subsection{LLM-enhanced Information Fusion}
The emergence of Large Language Models (LLMs) has introduced revolutionary paradigms for information fusion, transforming the integration and processing of heterogeneous data sources. Recent advancements demonstrate that LLMs serve as sophisticated fusion engines capable of harmonizing diverse information types and enabling semantic understanding across different data domains \cite{nguyen2024information}. Contemporary research has demonstrated several key approaches to LLM-based information fusion, including context-based fine-tuning frameworks that leverage information fusion principles to enhance model adaptability while preserving pre-trained knowledge \cite{nguyen2024information}. Novel techniques such as knowledge distillation and attention-based fusion mechanisms enable seamless integration of heterogeneous information sources into LLMs, representing significant advancement in intelligent data processing capabilities \cite{zhang2025distilling}. 
Various application domains have particularly benefited from LLM-enhanced information fusion approaches, where LLMs demonstrate exceptional capabilities in understanding, reasoning, and generation, introducing transformative paradigms for integrating artificial intelligence into complex decision-making systems \cite{xiao2024comprehensive}. These developments encompass fusion strategies operating at various architectural levels, from feature-level integration to decision-level fusion, with emerging hybrid approaches that leverage the representational power of LLMs to bridge semantic gaps effectively.
\subsection{LLMs in Healthcare Domains}
Large Language Models (LLMs), exemplified by ChatGPT \cite{ChatGPT}, demonstrate excellent performance in multidisciplinary applications.
Although current LLMs \cite{Baichuan,Baichuan4} with certain medical knowledge benefit from large-scale corpus support, they lack specialized medical proficiency and have significant performance bottlenecks in domain-specific scenarios.
Recently, several attempts~\cite{chen2023meditron70b,Chen2023HuatuoGPTIIOT, yang2024pediatricsgpt} have begun to build medically customized LLM assistants to fulfill the diagnostic and consultative demands.
For instance, HuatuoGPT series~\cite{Chen2023HuatuoGPTIIOT} have shown promising results in bridging generic-medical knowledge gaps by absorbing real doctor-patient conversations.
ZhongJing~\cite{yang2024zhongjing} improves Chinese medical capabilities by introducing expert feedback and multi-round medical instructions.
In addition, PediatricsGPT~\cite{yang2024pediatricsgpt} proposes a systematic training framework to construct interactive healthcare systems for pediatric specialists and medical generalists. Unlike previous studies, our framework aims to more fully recognize medical intents and refine the models' reasoning abilities in complex scenarios through the LLM-based multi-agent collaboration.

\subsection{LLM-based Multi-agent Collaboration}
With the focus of researchers on sophisticated goal-oriented dialog generation~\cite{fan2024aihospitalbenchmarkinglarge}, the inherent dilemmas of hallucinatory responses and weak comprehension in LLMs have been gradually exposed.
In this context, LLM-based automated agents are proposed to provide effective perception and decision-making skilfulness by incorporating external tools and databases~\cite{li2023camel}.
By mimicking human behavioral logic, multiple agents perform feedback and collaboration to enhance diverse intent understanding tasks, including educational training~\cite{wei2024medco} and emotional comfort~\cite{cheng2024cooper}.
For example, MEDCO~\cite{wei2024medco} enables LLMs to simulate interactions between patients and doctors, enhancing the practice performance of virtual students in interactive environments. MedAgents~\cite{tang2024medagentslargelanguagemodels} improves the performance of medical assistants in zero-shot settings through the role-playing strategy. MDAgents~\cite{kim2024mdagentsadaptivecollaborationllms} flexibly select the corresponding processing method according to the complexity of the inquiry. A detailed comparison between our framework and previous methods is provided in Figure ~\ref{demo}. In comparison, the proposed \textsc{MedAide} focuses more on mining profound healthcare intents and moving towards robust healthcare practices.

\section{Methodology}
Figure~\ref{arc} illustrates the overall architecture of the proposed \textsc{MedAide} framework. The workflow comprises three key components: Regularization-guided Information Extraction (RIE) module, dynamic Intent Prototype Matching (IPM) module, and Rotation Agent Collaboration (RAC) mechanism.

\begin{algorithm}[t]
\small
\caption{Syntactic Parsing Algorithm}
\label{alg:syntactic_parsing}
\begin{algorithmic}[noend]
\REQUIRE Input query $Q$
\ENSURE Parsed syntactic tree $T$
\STATE \textbf{Step 1: Tokenize the query}
\STATE Let $Q = { w_1, w_2, \dots, w_n }$ be the sequence of words (tokens) in the query.
\STATE \textbf{Step 2: Initialize parsing}
\STATE Initialize an empty syntactic tree $T = \emptyset$.
\STATE \textbf{Step 3: Apply CFG (Context-Free Grammar)}
\FOR {each non-terminal symbol $A$ in grammar $G$}
\STATE Find a rule $A \rightarrow \alpha$, where $\alpha$ is a string of terminals or non-terminals.
\IF {$A \rightarrow \alpha$ matches a substring of $Q$}
\STATE Add this production rule to the tree $T$.
\STATE Replace the matched string with non-terminal $A$.
\ENDIF
\ENDFOR
\STATE \textbf{Step 4: Recursion}
\WHILE {there are still non-terminals to be parsed}
\STATE Repeat step 3 until $T$ covers the entire query $Q$.
\ENDWHILE
\STATE \textbf{Step 5: Output the syntactic tree}
\RETURN $T$
\end{algorithmic}
\end{algorithm}

\begin{algorithm}[t]
\caption{LLM-Based Input Standardization}
\label{algorithm_input_std}
\small
\begin{algorithmic}
    \STATE \textbf{Input:} User query $Q_{input}$, Large Language Model $\mathcal{M}$, Set of rules $\mathcal{R}$ 
    \STATE \textbf{Output:} Standardized query $Q_{std}$
    \STATE \textbf{Initialize:} $Q_{cur} \gets Q_{input}$,\\ $converged \gets \text{False}$

    \WHILE{not $converged$}
        \STATE $converged \gets \text{True}$
        \FOR{each rule $r \in \mathcal{R}$}
            \STATE $Q_{new} \gets \mathcal{M}(Q_{cur}, r)$
            \IF{$Q_{new} \neq Q_{cur}$}
                \STATE $Q_{cur} \gets Q_{new}$
                \STATE $converged \gets \text{False}$
            \ENDIF
        \ENDFOR
    \ENDWHILE
    \STATE \textbf{return} $Q_{std} \gets Q_{cur}$
\end{algorithmic}
\normalsize
\end{algorithm}

\subsection{Regularization-guided Information Extraction}
Due to the inherent complexity and ambiguity of natural language expressions in medical contexts, the precise identification of complex medical intents and information extraction presents significant challenges~\cite{shaik2024survey}. Previous works have utilized rule-based approaches~\cite{aronson2010overview} and traditional natural language processing techniques to achieve certain levels of medical text understanding, but they suffer from the following limitations: (1) inadequate handling of composite medical intents that contain multiple interconnected clinical concerns; (2) insufficient semantic understanding of domain-specific medical terminology and contextual relationships. To address these challenges, we propose a Regularization-guided Information Extraction (RIE) module that leverages syntactic regularization algorithms combined with large language models to achieve robust medical query understanding and intent decomposition.

In this phase, we first process the initial query through the query input processor.
The processor is based on a set of syntactic regularization algorithms that combine LLMs with a predefined ruleset.
In this case, the core concept of RIE is to check, optimize, and normalize user input to $Q_{std}$. Algorithm~\ref{alg:syntactic_parsing} shows the corresponding program.
First, the algorithm disambiguates the input query and divides the query into individual words or tokens. Next, the algorithm initializes an empty syntax tree and matches substrings in the query by context-free grammar rules, adding the conforming rules to the syntax tree. The algorithm keeps applying the rules recursively until the syntax tree covers the entire query and finally outputs the complete syntax tree.

\begin{figure*}[t]
  \centering
  \includegraphics[width=1.0\linewidth]{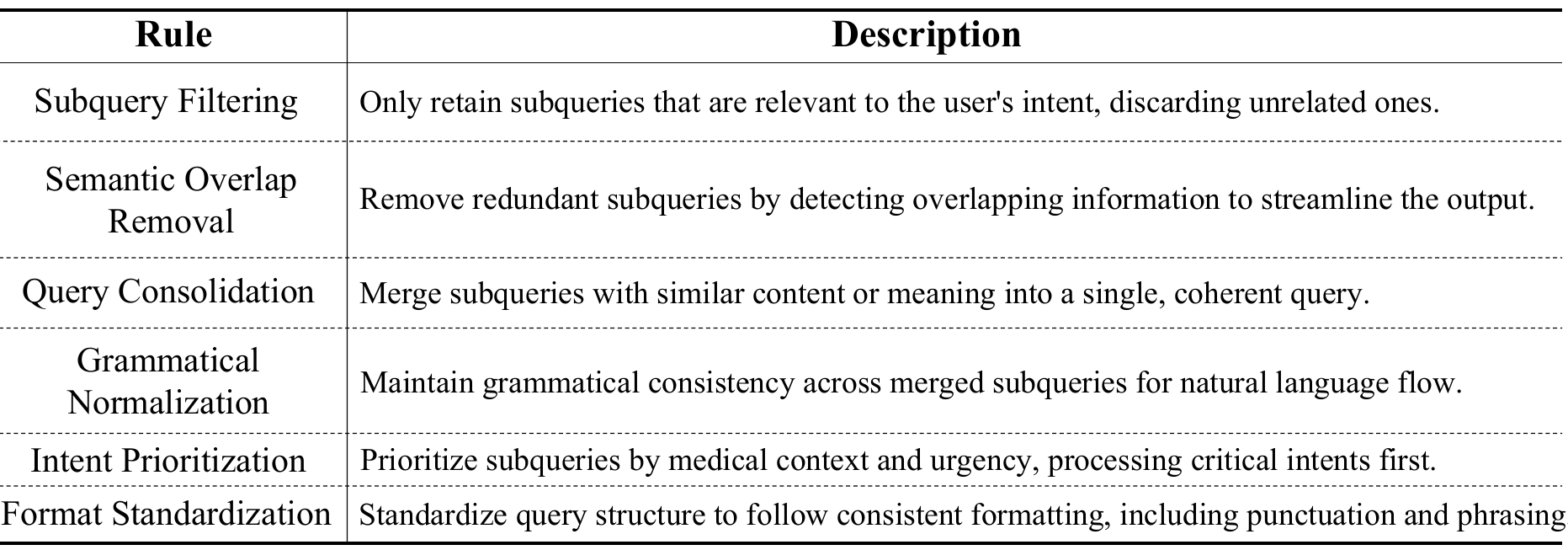}
  \caption{Illustration of the rule sets.
  }
  \label{rule_sets}
  \vspace{-5pt}
\end{figure*}

Then, a key element extractor distils critical information from $Q_{std}$, such as symptoms, condition descriptions, and medical histories, to form the element set $E_{i}$. In the Retrieval-Augmented Generation (RAG), we build an indexed database of 1,095 expert-reviewed medical guidelines and retrieve documents related to these elements using a semantic retrieval method ~\cite{lewis2021retrieval}, forming the document set $D_{ref}$.
Subsequently, the recall analysis module identifies relevant documents through prompt-guided selection based on BERT-Score \cite{zhang2020bertscoreevaluatingtextgeneration}, which is fed into the LLM-based query prompter along with $Q_{std}$.
The prompter optimizes the empirical information and efficient decomposition of composite intents. Eventually, we design a refined query constructor to merge and integrate the generated subqueries.
To ensure that the generated subqueries maintain high quality in both semantic meaning and formal structure, we implement a set of rules (as shown in Figure~\ref{rule_sets}) in the refined query constructor using Algorithm~\ref{algorithm_input_std}. These rules encompass subquery filtering, semantic overlap removal, query consolidation, grammatical normalization, intent prioritization, and format standardization. Specifically, the constructor filters and reconstructs multiple subqueries $Q_{gen}$ with these predefined rulesets to ensure semantically complete and formally uniform output. These processes streamline the output, optimize the query structure, and prioritize critical medical intents, ensuring that the final query is semantically coherent and structurally consistent.

\begin{table*}[t]
\centering
\renewcommand{\arraystretch}{1.2}
\setlength{\tabcolsep}{4pt}
\caption{Comparison results on the Pre-Diagnosis benchmark.}
\resizebox{1\linewidth}{!}{%
\begin{tabular}{l|cccccc}
\toprule
\multicolumn{1}{c|}{{\textbf{Models}}}         & { BLEU-1}         & { BLEU-2}       & { Meteor}         & { BERT-Score}        & { ROUGE-L}         & { GLEU}      \\ \midrule
{ZhongJing/ + \textsc{MedAide}}    & { 10.24/\textbf{12.44}}   & { 3.74/\textbf{4.56}} & { 14.51/\textbf{15.78}} & { 47.54/\textbf{48.06}}   & { 11.12/\textbf{12.31}} & { 3.87/\textbf{4.69}} \\
{Meditron-7B/ + \textsc{MedAide}}    & { 3.76/\textbf{4.94}}   & { 0.91/\textbf{1.30}} & { 6.63/\textbf{8.51}}  & { 42.64/\textbf{42.75}}  & { 4.52/\textbf{5.21}}   & { 1.74/\textbf{2.07}} \\
{HuatuoGPT-II/ + \textsc{MedAide}} & { 14.6/\textbf{15.18}}  & { 5.87/\textbf{6.63}} & { 18.81/\textbf{21.84}} & { 49.26/\textbf{51.88}}  & { 13.57/\textbf{19.04}} & { 6.43/\textbf{7.25}} \\ 
{Baichuan4/ + \textsc{MedAide}}      & {12.97/\textbf{15.95}} & { 5.39/\textbf{7.85}} & { 16.68/\textbf{21.60}} & { 48.68/\textbf{52.69}} & { 13.33/\textbf{20.75}} & { 5.57/\textbf{7.98}} \\ \midrule
{LLama-3.1-8B/ + \textsc{MedAide}}   & { 10.95/\textbf{15.36}} & { 2.13/\textbf{6.78}} & { 15.47/\textbf{19.54}} & { 47.62/\textbf{51.41}}  & { 11.45/\textbf{19.29}} & { 4.37/\textbf{6.45}} \\
{GPT-4o/ + \textsc{MedAide}}         & { 15.28/\textbf{15.93}} & { 6.33/\textbf{7.56}} & { 19.89/\textbf{21.78}} & {51.95/\textbf{53.65}}  & { 14.15/\textbf{17.64}} & {6.34/\textbf{7.31}} \\
{Claude 3.7 Sonnet/ + \textsc{MedAide}}         & { 15.67/\textbf{16.42}} & { 6.55/\textbf{7.89}} & { 20.31/\textbf{22.47}} & {52.34/\textbf{57.88}}  & { 14.62/\textbf{18.21}} & {6.51/\textbf{7.65}} \\
{DeepSeek-R1/ + \textsc{MedAide}}         & { 16.03/\textbf{16.78}} & { 6.82/\textbf{8.14}} & { 20.85/\textbf{23.15}} & {52.86/\textbf{54.92}}  & { 15.04/\textbf{18.73}} & {6.73/\textbf{7.93}} 
\\ \bottomrule
\end{tabular}
}
\label{PreDiagnosis}
\end{table*}

\begin{table*}[t]
\centering
\renewcommand{\arraystretch}{1.2}
\setlength{\tabcolsep}{4pt}
\caption{Comparison results on the Diagnosis benchmark.}
\resizebox{1\linewidth}{!}{%
\begin{tabular}{l|cccccc}
\toprule
\multicolumn{1}{c|}{{\textbf{Models}}}         & { BLEU-1}         & { BLEU-2}       & { Meteor}         & { BERT-Score}        & { ROUGE-L}         & { GLEU}      \\ \midrule
{ZhongJing/ + \textsc{MedAide}}             & { 11.21/\textbf{12.33}} & { 5.12/\textbf{6.25}}   & { 26.05/\textbf{27.13}} & { 5.23/\textbf{5.67}}   & { 11.34/10.87} & { 6.14/5.49}  \\
{Meditron-7B/ + \textsc{MedAide}}             & { 12.01/\textbf{13.93} } & { 3.13/\textbf{6.03} }   & { 19.10/\textbf{26.53} } & { 3.60/\textbf{6.12} }   & { 8.60/\textbf{11.67} }  & { 3.99/\textbf{6.48} }  \\
{HuatuoGPT-II/ + \textsc{MedAide}}          & { 22.65/\textbf{25.26}} & {10.21/\textbf{12.11}} & { 34.51/\textbf{40.32}} & { 8.36/\textbf{11.73}}  & { 15.77/\textbf{19.11}} & { 8.95/\textbf{10.65}} \\ 
{Baichuan4/ + \textsc{MedAide}}               & { 22.58/\textbf{25.51}} & {10.98/\textbf{13.61}} & { 36.26/\textbf{42.77}} & { 10.30/\textbf{15.00}} & { 17.03/\textbf{22.43}} & { 9.61/\textbf{11.95}} \\ \midrule
{LLama-3.1-8B/ + \textsc{MedAide}}            & { 16.92/\textbf{22.75}} & {6.56/\textbf{9.97}}   & { 26.39/\textbf{38.52}} & { 4.78/\textbf{10.41}}  & {12.79/\textbf{18.68}} & { 6.21/\textbf{8.94}}  \\
{GPT-4o/ + \textsc{MedAide}}                  & { 21.67/\textbf{26.80}} & {9.73/\textbf{14.56}}  & { 32.67/\textbf{44.11}} & { 7.95/\textbf{15.81}}  & { 16.20/\textbf{23.67}} & { 8.41/\textbf{12.69}} \\ 
{Claude 3.7 Sonnet/ + \textsc{MedAide}}         & { 22.14/\textbf{27.41}} & {10.36/\textbf{15.02}} & { 33.85/\textbf{45.23}} & {8.64/\textbf{16.32}}  & { 16.78/\textbf{24.14}} & {8.92/\textbf{13.08}} \\
{DeepSeek-R1/ + \textsc{MedAide}}         & { 22.83/\textbf{28.16}} & {10.89/\textbf{15.67}} & { 34.91/\textbf{46.54}} & {9.23/\textbf{17.05}}  & { 17.45/\textbf{24.88}} & {9.38/\textbf{13.52}} 
\\
\bottomrule
\end{tabular}
}
\label{Diagnosis}
\end{table*}

\subsection{Dynamic Intent Prototype Matching}
Medical intent recognition in healthcare systems faces significant challenges due to the diverse expressions of similar clinical concerns and the subtle semantic differences between related medical intents~\cite{li2024deep,muhammad2022comprehensive}. Previous approaches for intent classification have employed traditional machine learning methods~\cite{zhang2016joint}, achieving reasonable performance in general domains but exhibiting limitations in medical contexts: (1) insufficient capture of domain-specific semantic nuances in medical terminology and clinical expressions; (2) inadequate handling of fine-grained intent distinctions that are crucial for accurate medical decision-making~\cite{yang2016hierarchical}. To address these challenges, we propose a dynamic Intent Prototype Matching (IPM) module that leverages contextualized medical embeddings to achieve robust semantic understanding and precise intent classification in healthcare applications.

After the RIE module, the optimized query $Q_{opt}$ is matched with a set of intent prototype embeddings $E_i$ generated by a contextual encoder, which is designed to capture the semantic features of different medical intents. 
The encoder is constructed on top of BioBERT~\cite{Lee2019BioBERTAP} to learn prototype representations by performing fine-grained intent classification.
We add a fully connected layer after the embedding layer with output dimensions aligned to the 17 medical intent categories and generate the corresponding probability distributions via a softmax activation function.
Specifically, the contextual encoder maps the optimized query together with the intent embeddings into a 768-dimensional embedding space. It computes the cosine similarity $S_{ij}$ between the query and each intent embedding $E_i$ with the following formula:
\begin{equation}
    S_{ij} = \frac{Q_{opt} \cdot E_{i}}{\left\| Q_{opt} \right\| \left\| E_{i} \right\|}.
    \label{eq1}
\end{equation}

Subsequently, the probability distribution $\alpha_{ij}$ for each intent after the softmax is expressed as:
\begin{equation}
    \alpha_{ij} = \frac{\exp(S_{ij})}{\sum_{l=1}^{17} \exp(S_{il})}.
    \label{eq2}
\end{equation}
If the probability $ \alpha _{ij} $ of an intent $ i $ exceeds a predetermined threshold, the intent will be activated, triggering the corresponding agent:
\begin{equation}
    \text{Activated Intent}_{i} = 
    \left\{
    \begin{array}{ll}
    1 & \text{if } \alpha_{ij} > \text{Threshold}, \\
    0 & \text{otherwise}.
    \end{array}
    \right.
\end{equation}
In this way, our framework can automatically activate the most compliant medical intent based on the optimized query, directing it to the corresponding agent to perform subsequent operations.

\subsection{Rotation Agent Collaboration}

Effective multi-agent collaboration in healthcare systems requires sophisticated coordination mechanisms to ensure coherent and comprehensive patient care. Traditional multi-agent approaches often suffer from information fragmentation and a lack of systematic integration across different medical specialties~\cite{du2021survey}. We propose a Rotation Agent Collaboration (RAC) mechanism that systematically coordinates agents through sequential leadership transitions, ensuring comprehensive coverage of healthcare services from pre-diagnosis to post-treatment.

The multi-agent collaboration framework operates through the dynamic activation of specialized agents based on IPM. Our polling-based information fusion mechanism enables systematic knowledge integration across four distinct medical domains: pre-diagnosis assessment, diagnostic reasoning, medication management, and post-treatment care. Each domain has a designated primary contact agent that coordinates the fusion process for domain-specific decision-making.

\noindent\textbf{Pre-Diagnosis Assessment with Multi-Source Fusion.}
When pre-diagnosis intents are detected, the pre-diagnosis agent serves as the primary contact for orchestrating the synthesis of patient information. The agent maintains a relational database repository~\cite{codd1970relational} and implements cross-agent polling to fuse comprehensive patient profiles. The fusion mechanism integrates the diagnostic agent output for the preliminary interpretation of symptoms with the analysis of the medication agent's patterns of medication history. This multi-source approach enables enhanced risk stratification and informed departmental routing decisions through the systematic integration of knowledge.

\noindent\textbf{Diagnostic Reasoning through Hybrid Knowledge Fusion.}
For diagnostic intents, the diagnosis agent serves as the main contact, coordinating information fusion across 506 high-quality medical cases from the medical record database~\cite{fan2024aihospitalbenchmarkinglarge}. The agent employs a hybrid retrieval-fusion mechanism~\cite{sawarkar2024blendedragimprovingrag} enhanced by cross-agent knowledge integration. The polling process incorporates pre-diagnosis agent outputs for patient historical context and medicament agent analyses for drug-symptom interaction patterns.

The hybrid information retrieval operates through parallel keyword and semantic fusion channels. The keyword-based channel extracts document subsets $D_{\text{slice}}$ through lexical matching:
\begin{equation}
D_{\text{slice}} = \{ d \in D \mid \text{KeywordMatch}(Q, d) = True \}.
\end{equation}

The semantic channel computes similarity scores $S(Q, d)$ between queries and documents using embedding-based fusion:
\begin{equation}
S(Q, d) = \frac{E_Q \cdot E_D}{\| E_Q \| \| E_D \|},
\end{equation}
\begin{equation}
D_{\text{match}} = \{ d \in D \mid S(Q, d) > \tau \}.
\end{equation}

The final information fusion combines both retrieval channels:
\begin{equation}
D_{\text{final}} = D_{\text{slice}} \cup D_{\text{match}}.
\end{equation}

\noindent\textbf{Medication Management via Cross-Agent Integration.}
When medicament intents are activated, the medicament agent coordinates knowledge fusion across 26,684 medication entries from PubMed~\cite{fiorini2018best}. The polling-based integration mechanism fuses diagnostic agent outputs for confirmed conditions with pre-diagnosis agent data regarding allergy histories and contraindications. This multi-source fusion enables dynamic dosage optimization and interaction detection tailored to individual patient profiles through systematic information synthesis.

\noindent\textbf{Post-Treatment care through Clinical Data Integration.}
For post-diagnosis intents involving care support and rehabilitation advice, the post-diagnosis agent coordinates information synthesis from clinical decision outputs. The integration process fuses diagnostic agent prognosis data with medicament agent adherence to develop personalized rehabilitation protocols and recovery milestone frameworks. This approach ensures continuity between active treatment and long-term care management through comprehensive data fusion.

\noindent\textbf{Polling-Based Coordination Protocol.}
The polling mechanism ensures systematic information flow between main contact transitions. For each stage $s \in \{1, 2, 3, 4\}$, the main contact agent $A_s^{mc}$ coordinates with supporting agents $\{A_1, A_2, \dots, A_n\} \setminus A_s^{mc}$:
\begin{equation}
O_s^{(0)} = A_s^{mc}(P_s, D_{\text{input}}^{(s)}),
\end{equation}
where $P_s$ represents the stage-specific prompt and $D_{\text{input}}^{(s)}$ includes both patient data and outputs from previous stages. Each supporting agent contributes specialized knowledge:
\begin{equation}
C_i^{(s)} = A_i(P_s, D_{\text{input}}^{(s)}, O_s^{(0)}), \quad i \neq s,
\end{equation}
The main contact agent then integrates all contributions to produce the stage output, and the final comprehensive output combines all stage results:
\begin{equation}
O_s^{final} = \text{Integrate}_s(O_s^{(0)}, \{C_i^{(s)} \mid i \neq s\}),
\end{equation}
\begin{equation}
\text{Output}_{\text{final}} = \text{Synthesize}(O_1^{final}, O_2^{final}, O_3^{final}, O_4^{final}).
\end{equation}
This polling-based approach ensures that each healthcare stage receives focused attention from the most qualified agent while maintaining systematic integration across the entire care continuum.

\begin{figure*}[t]
  \centering
  \includegraphics[width=1\linewidth]{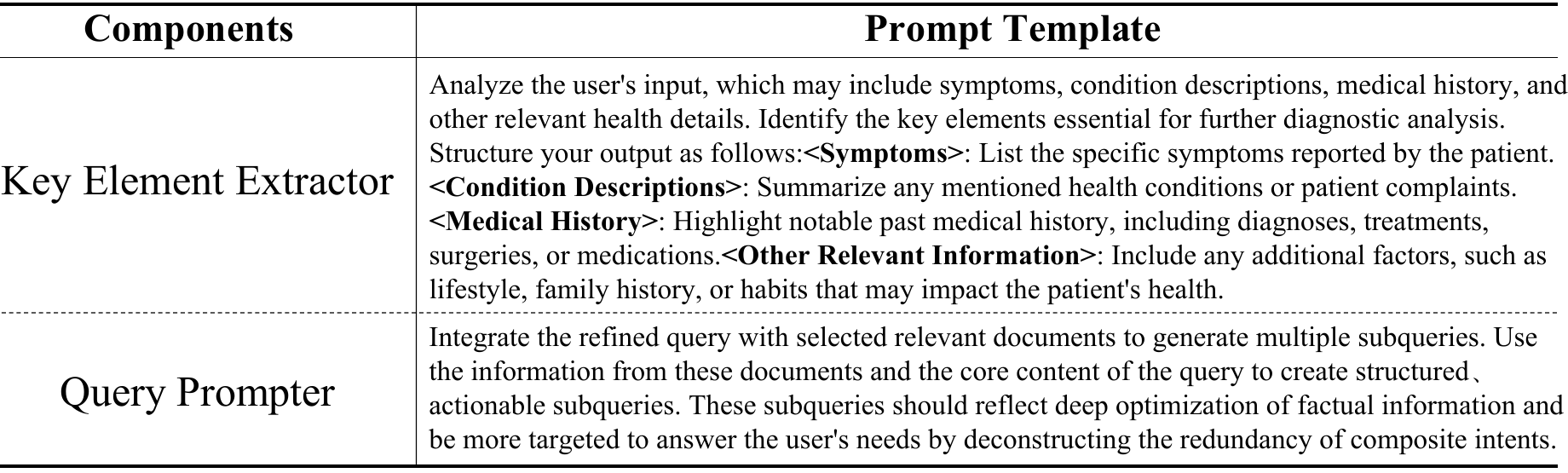}
  \caption{Illustration of the prompt template for the RIE.
  }
  \label{app:2_1}
\end{figure*}

\begin{figure*}[t]
  \centering
  \includegraphics[width=1\linewidth]{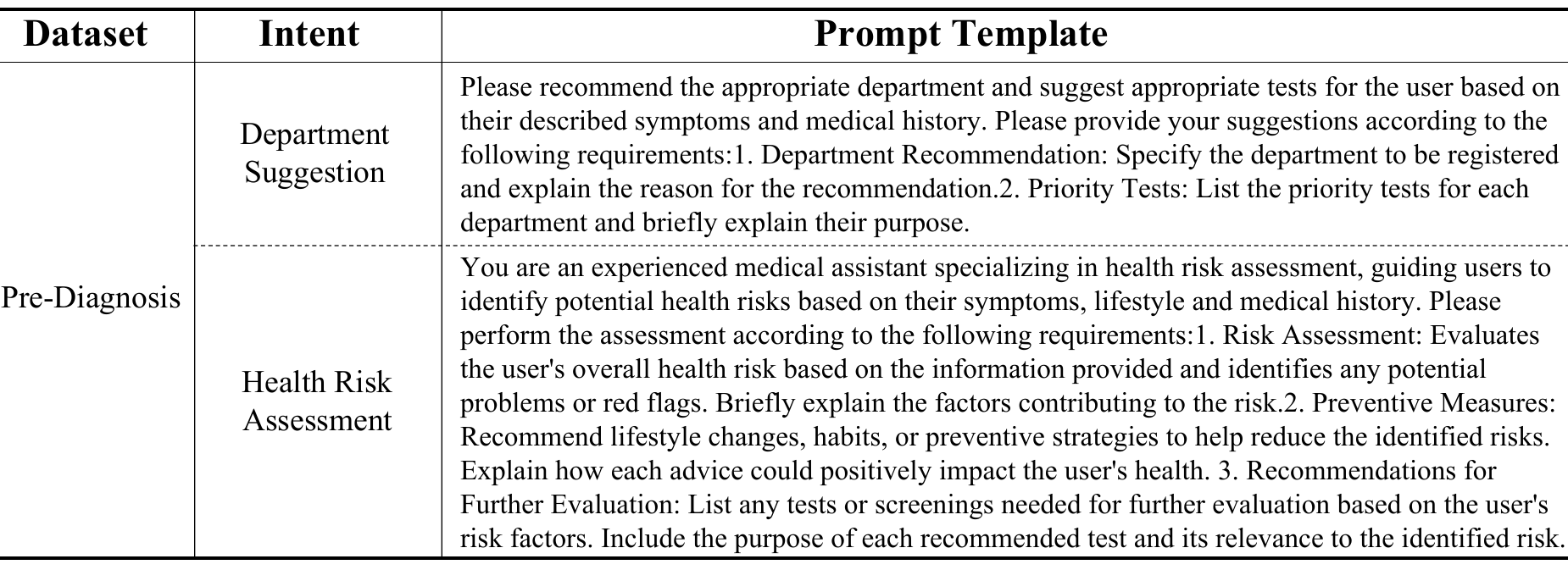}
  \caption{Illustration of the prompt template for the Pre-Diagnosis.
  }
  \label{app:2_2}
\end{figure*}

\begin{figure*}[t]
  \centering
  \includegraphics[width=1\linewidth]{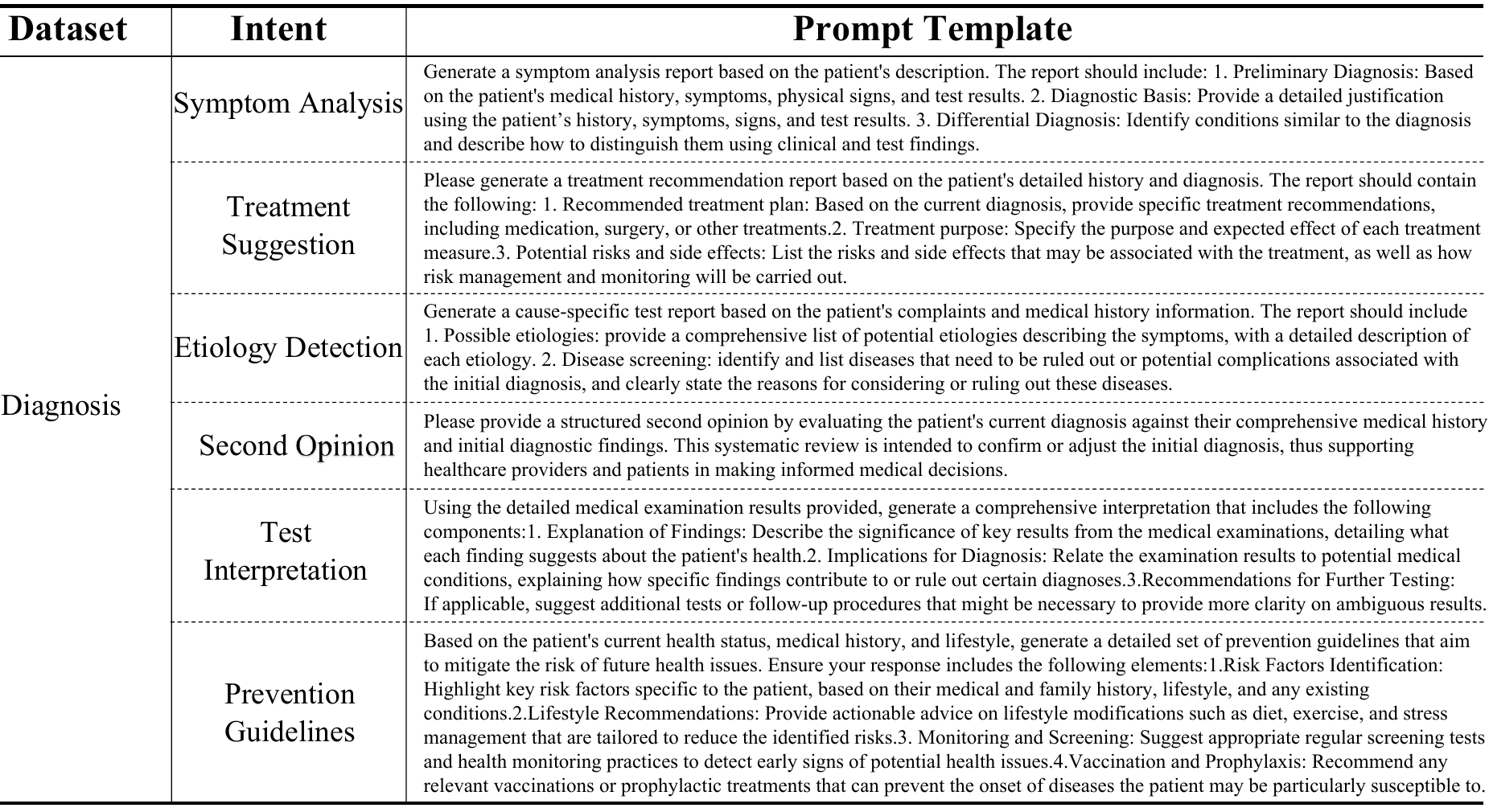}
  \caption{Illustration of the prompt template for the Diagnosis.
  }
  \label{app:2_3}
\end{figure*}

\begin{figure*}[t]
  \centering
  \includegraphics[width=1.0\linewidth]{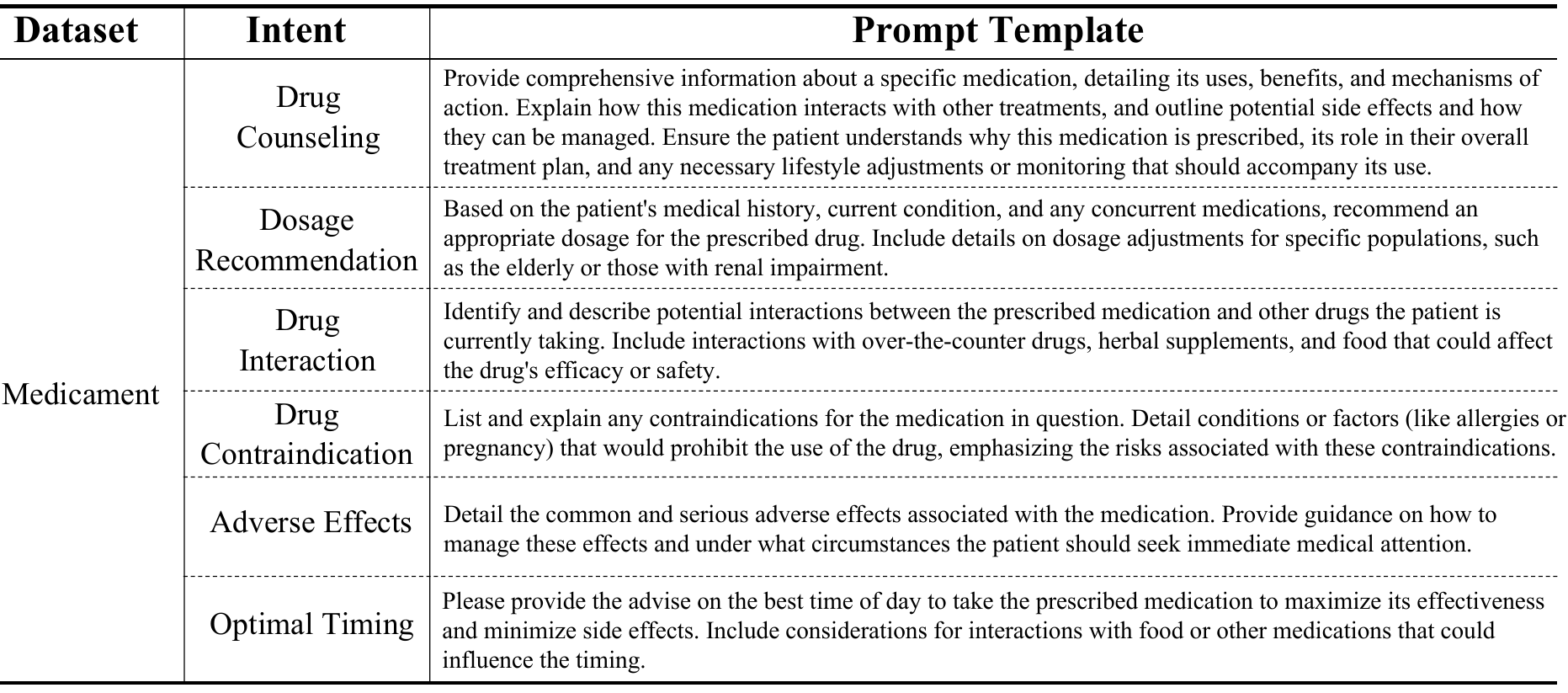}
  \caption{Illustration of the prompt template for the Medicament.
  }
  \label{app:2_4}
\end{figure*}

\begin{figure*}[t]
  \centering
  \includegraphics[width=1.0\linewidth]{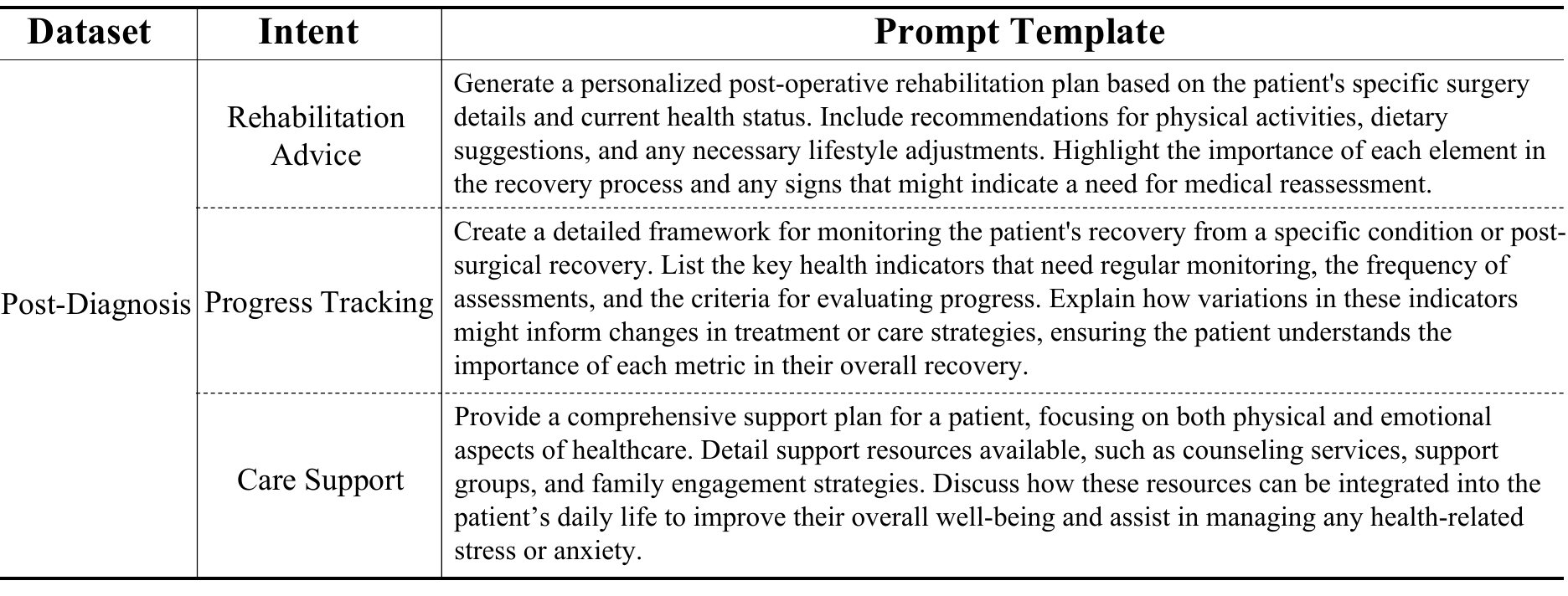}
  \caption{Illustration of the prompt template for the Post-Diagnosis.
  }
  \label{app:2_5}
\end{figure*}

\section{Experiments}

\subsection{Datasets and Implementation Details}
All experiments are conducted in the zero-shot setting.
To reflect the real-world demands for healthcare services, we conduct four different benchmarks across 17 types of medical intents, including Pre-Diagnosis, Diagnosis, Medicament, and Post-Diagnosis benchmarks.
Each benchmark contains 500 composite intent instances.
We employ structured prompt templates across four distinct agents within our framework. Each prompt is designed to guide the language model in generating contextually relevant and accurate responses tailored to specific medical tasks. These templates are instrumental in ensuring that the interactions between the agents and users are effective and medically informative. The prompt templates for the various components of the RIE phase are detailed in Figure~\ref{app:2_1}. The specifics for each agent's prompt templates are depicted in Figures~\ref{app:2_2}, \ref{app:2_3}, \ref{app:2_4}, and \ref{app:2_5}.

To enable fine-grained intent prototype matching, we develop a contextual encoder training dataset consisting of 2,800 training samples, 300 validation samples, and 300 testing samples. For the medical intent classification, we use the BioBERT-based contextual encoder from the HuggingFace library\footnote{\url{https://huggingface.co/dmis-lab/biobert-v1.1}} to map the input sentences to a 768-dimensional embedding space. For each intent classification task, we add a fully connected layer after the embedding layer with an output dimension of 17 and generate probability distributions corresponding to the 17 intents via a softmax activation function. To handle these tasks, we deploy four V100 GPUs (32 GB VRAM each) to provide sufficient computational resources and graphics memory. The Adam optimizer is used in the training process with an initial learning rate of 1e-6 and 5 training rounds, and a learning rate scheduling strategy is applied in the validation set for dynamic adjustment. The loss function is the cross-entropy loss function. We evaluate the model using accuracy and F1 score to ensure its performance on the validation set.

\begin{table*}[t]
\centering
\renewcommand{\arraystretch}{1.2}
\setlength{\tabcolsep}{4pt}
\caption{Comparison results on the Medicament benchmark.}
\vspace{2pt}
\resizebox{1.0\linewidth}{!}{%
\begin{tabular}{l|cccccc}
\toprule
\multicolumn{1}{c|}{{\textbf{Models}}}         & {BLEU-1}         & { BLEU-2}       & {Meteor}         & {BERT-Score}        & {ROUGE-L}         & {GLEU}      \\ \midrule
{ZhongJing/ + \textsc{MedAide}}              & {13.14/\textbf{13.87}} & {5.02/\textbf{5.34}}  & {22.68/\textbf{25.09}} & {5.48/\textbf{5.99}}   & {14.01/\textbf{14.85}} & {6.02/\textbf{6.43}}  \\
{Meditron-7B/ + \textsc{MedAide}}              & {5.76/\textbf{19.43}}  & {2.04/\textbf{17.50}} & {12.29/\textbf{34.84}} & {2.12/\textbf{29.43}}  & {6.31/\textbf{32.57}}  & {2.79/\textbf{24.47}} \\
{HuatuoGPT-II/ + \textsc{MedAide}}           & {17.14/\textbf{37.89}} & {9.61/\textbf{27.47}} & {30.72/\textbf{58.06}} & {11.96/\textbf{33.62}} & {19.96/\textbf{45.39}} & {9.57/\textbf{24.23}} \\ 
{Baichuan4/ + \textsc{MedAide}}                & {14.64/\textbf{44.63}} & {7.12/\textbf{36.34}} & {26.26/\textbf{62.11}} & {8.09/\textbf{42.56}}  & {15.47/\textbf{49.91}} & {7.08/\textbf{35.99}} \\ \midrule
{LLama-3.1-8B/ + \textsc{MedAide}}             & {13.80/\textbf{23.87}} & {5.90/\textbf{16.75}} & {27.28/\textbf{47.39}} & {7.50/\textbf{26.75}}  & {16.05/\textbf{38.30}} & {6.37/\textbf{15.77}} \\
{GPT-4o/ + \textsc{MedAide}}                   & {16.23/\textbf{45.13}} & {8.16/\textbf{38.96}} & {29.24/\textbf{65.09}} & {9.78/\textbf{40.53}}  & {17.53/\textbf{52.71}} & {8.61/\textbf{37.09}} \\ 
{Claude 3.7 Sonnet/ + \textsc{MedAide}}         & {16.88/\textbf{46.24}} & {8.67/\textbf{39.85}} & {30.41/\textbf{66.27}} & {10.52/\textbf{41.86}}  & {18.24/\textbf{53.94}} & {9.15/\textbf{38.16}} \\
{DeepSeek-R1/ + \textsc{MedAide}}         & {17.42/\textbf{47.38}} & {9.21/\textbf{40.92}} & {31.63/\textbf{67.83}} & {11.27/\textbf{43.15}}  & {19.06/\textbf{55.18}} & {9.74/\textbf{39.28}} 
\\
\bottomrule
\end{tabular}
}
\label{medicament}
\end{table*}

\begin{table*}[t]
\centering
\renewcommand{\arraystretch}{1.2}
\setlength{\tabcolsep}{4pt}
\caption{Comparison results on the Post-Diagnosis benchmark.}
\resizebox{1.0\linewidth}{!}{%
\begin{tabular}{l|cccccc}
\toprule
\multicolumn{1}{c|}{{\textbf{Models}}}         & { BLEU-1}         & { BLEU-2}       & { Meteor}         & { BERT-Score}        & {ROUGE-L}         & { GLEU}      \\ \midrule
{ZhongJing/ + \textsc{MedAide}}                   & { 13.03/\textbf{17.83}} & { 5.08/\textbf{5.94}}   & { 26.22/\textbf{32.14}} & { 4.23/\textbf{5.22}}   & { 13.48/\textbf{15.34}} & { 6.21/\textbf{7.59}}   \\
{Meditron-7B/ + \textsc{MedAide}}                   & { 8.72/\textbf{12.78}}  & { 2.46/\textbf{3.06}}   & { 18.75/\textbf{20.21}} & { 1.77/\textbf{4.32}}   & { 7.28/\textbf{8.58}}   & {3.84/\textbf{4.40}}   \\
{HuatuoGPT-II/ + \textsc{MedAide}}                & { 21.39/\textbf{26.10}} & { 12.24/\textbf{12.46}} & { 35.55/\textbf{42.23}} & { 11.02/\textbf{13.62} } & { 13.62/\textbf{21.18}} & { 10.59/\textbf{10.96} } \\
{Baichuan4/ + \textsc{MedAide}}                     & { 15.89/\textbf{26.10}} & { 7.06/\textbf{13.92}}  & { 27.69/\textbf{44.55}} & { 6.92/\textbf{13.05}}  & { 14.86/\textbf{21.35}} & { 6.59/\textbf{11.61}}  \\ \midrule
{LLama-3.1-8B/ + \textsc{MedAide}}                  & { 21.85/\textbf{22.95}} & { 7.42/\textbf{7.68}}   & { 35.36/\textbf{37.71}} & { 5.22/\textbf{7.97}}   & { 15.02/\textbf{17.93}} & { 7.65/\textbf{8.01}}   \\
{GPT-4o/ + \textsc{MedAide}}                        & { 19.27/\textbf{26.83}} & { 9.28/\textbf{11.54}}  & { 30.72/\textbf{40.83}} & { 8.64/\textbf{8.71}}   & { 16.22/\textbf{18.12}} & { 8.41/\textbf{9.89}}   \\ 
{Claude 3.7 Sonnet/ + \textsc{MedAide}}         & { 20.14/\textbf{27.52}} & { 9.87/\textbf{12.05}} & { 32.25/\textbf{42.16}} & {9.12/\textbf{9.83}}  & { 16.84/\textbf{18.87}} & {8.92/\textbf{10.43}} \\
{DeepSeek-R1/ + \textsc{MedAide}}         & { 20.92/\textbf{28.27}} & { 10.35/\textbf{12.63}} & { 33.78/\textbf{43.52}} & {9.75/\textbf{9.86}}  & { 17.51/\textbf{19.58}} & {9.43/\textbf{10.94}} 
\\
\bottomrule
\end{tabular}
}
\label{PostDiagnosis}
\end{table*}

\begin{figure*}[t]
  \centering
  \includegraphics[width=\linewidth]{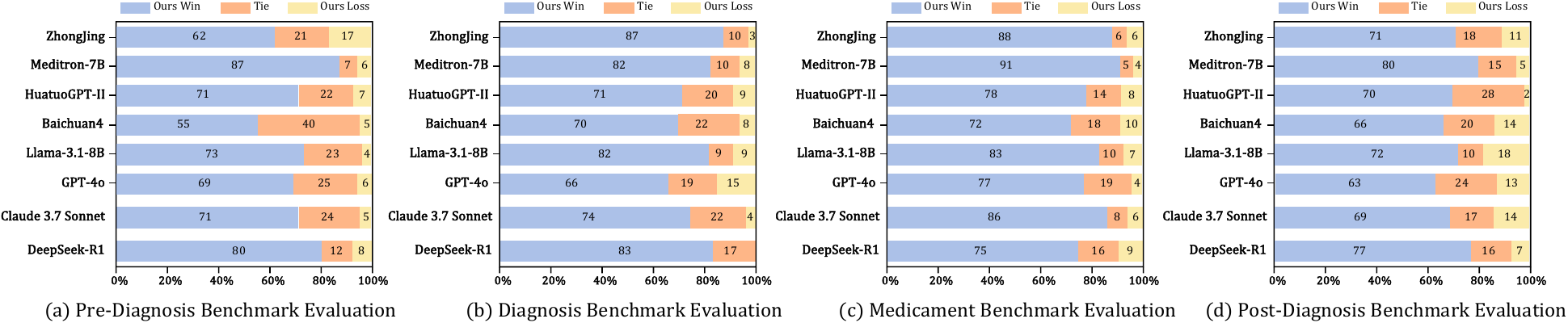}
  \caption{Response comparisons of \textsc{MedAide} with other baselines via GPT-4o evaluation.
  }
  \label{vis}
\end{figure*}

\subsection{Model Zoo}
We compare a series of state-of-the-art (SOTA) models for comprehensive evaluations. Among medical LLMs, \textbf{ZhongJing}~\cite{yang2024zhongjing} is a traditional Chinese medicine model based on the Ziya-LLaMA model \cite{ziya-llama} with complete training procedures.
\textbf{Meditron-7B} \cite{chen2023meditron70b} performs medically relevant continuous pre-training on Llama-2-7B to extend the breadth and depth of the model in medical knowledge. \textbf{HuatuoGPT-II (13B)} \cite{Chen2023HuatuoGPTIIOT} employs a one-stage unified approach for domain adaptation to improve medical expertise.
\textbf{Baichuan4} \cite{Baichuan4} optimizes long texts by scaling law and combines reinforcement learning techniques to improve reasoning ability with enhanced medical knowledge representation.
In general-purpose models, \textbf{Llama-3.1-8B} \cite{dubey2024llama} relies on the grouped query attention mechanism to enhance the inference efficiency and performs well on multilingual tasks.
\textbf{GPT-4o} \cite{openai2024gpt4ocard} shows excellent language comprehension and generation capabilities and excels in handling complex tasks. \textbf{Claude 3.7 Sonnet} \cite{anthropic2025claude} is designed for advanced reasoning with balanced performance and speed, featuring enhanced contextual understanding capabilities. \textbf{DeepSeek-R1}~\cite{deepseekai2025deepseekr1incentivizingreasoningcapability} incorporates advanced multi-stage reasoning techniques with recursive refinement, demonstrating superior performance on complex reasoning tasks and knowledge integration.

\subsection{Comparison with SOTA Methods}
As a plug-and-play framework, we combine \textsc{MedAide} with the baseline models to provide comprehensive evaluations by different metrics, including BLEU-1/2 (\%)~\cite{papineni-etal-2002-bleu}, ROUGE-1/2/L (\%)~\cite{lin-2004-rouge}, GLEU (\%)~\cite{zhang2020bertscoreevaluatingtextgeneration}, Meteor (\%)~\cite{zhang2020bertscoreevaluatingtextgeneration}, BERT-Score (\%)~\cite{wu2016googles}.

\begin{figure*}[t]
  \centering
  \includegraphics[width=\linewidth]{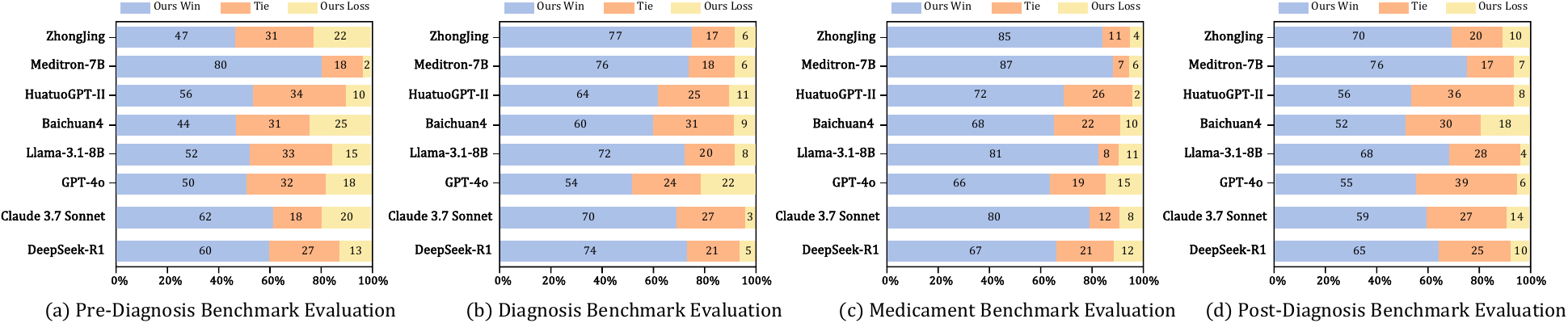}
  \caption{Response comparisons of \textsc{MedAide} with other baselines via doctor evaluation.
  }
  \label{vis}
\end{figure*}

\begin{table*}[t]
\centering
\caption{Ablation study results demonstrating component effectiveness across four medical benchmarks. w/'' and w/o'' are short for with and without, respectively.}
\resizebox{1.0\linewidth}{!}{%
\begin{tabular}{ccccccccc}
\toprule
{}                             & \multicolumn{2}{c}{{\textbf{Pre-Diagnosis}}}                                            & \multicolumn{2}{c}{{\textbf{Diagnosis}}}                                                 & \multicolumn{2}{c}{{\textbf{Medicament}}}                                                & \multicolumn{2}{c}{{\textbf{Post-Diagnosis}}}                                           \\ \cline{2-9} 
\multirow{-2}{*}{{\textbf{Components}}} & \multicolumn{1}{l}{{ROUGE-L}} & \multicolumn{1}{l}{{GLEU}} & \multicolumn{1}{l}{{ROUGE-L}} & \multicolumn{1}{l}{{GLEU}} & \multicolumn{1}{l}{{ROUGE-L}} & \multicolumn{1}{l}{{GLEU}} & \multicolumn{1}{l}{{ROUGE-L}} & \multicolumn{1}{l}{{GLEU}} \\ \hline
{Full \textsc{MedAide}}                   & {\textbf{17.64}}              & {\textbf{7.31}}            & {\textbf{23.67}}              & {\textbf{12.69}}           & {\textbf{52.71}}              & {\textbf{37.09}}           & {\textbf{18.12}}              & {\textbf{9.89}}            \\ \midrule
{w/o RIE}          & {16.02}                       & {6.44}                     & {18.35}                       & {10.37}                    & {32.06}                       & {32.45}                    & {16.48}                       & {8.79}                     \\
{w/ GPT-4o Recognition}         & {16.58}                       & {6.73}                     & {22.98}                       & {12.34}                    & {44.72}                       & {34.10}                    & {17.56}                       & {9.02}                     \\
{w/o Decision Analysis}        & {17.22}                       & {7.22}                     & {23.14}                       & {12.54}                    & {48.72}                       & {35.25}                    & {18.04}                       & {9.56}                     \\ \bottomrule
\end{tabular}
}
\label{Ablation}
\end{table*}

\noindent \textbf{Results on Pre-Diagnosis Benchmark.}
As Table~\ref{PreDiagnosis} shows, our framework consistently improves the performance across metrics for all models.
(i) With \textsc{MedAide} support, HuatuoGPT-II and Baichuan4 achieve significant relative improvements in BERT score, indicating stronger diagnostic semantic alignment.
(ii) Additionally, larger frontier models like DeepSeek-R1 and Claude 3.7 Sonnet show substantial improvements with \textsc{MedAide}, with DeepSeek-R1 showing an 11.0\% increase in Meteor scores and Claude 3.7 Sonnet achieving a 10.6\% improvement in BERT-Score, highlighting the framework's effectiveness even with SOTA models.
(iii) Furthermore, \textsc{MedAide} helps general LLama-3.1-8B to improve content quality with a notable Meteor score increase from 15.47\% to 19.54\%, demonstrating better precision in medical terminology and symptom descriptions.
(iv) Despite the performance constraints of ZhongJing and Meditron-7B in tackling complex medical scenarios due to the scaling law, our framework still provides performance gains across all evaluated metrics.

\noindent \textbf{Results on Diagnosis Benchmark.}
Table~\ref{Diagnosis} shows the diagnosis task results containing six composite intent scenarios.
(i) Combined with our framework, Meditron-7B and HuatuoGPT-II outperform vanilla baselines in response accuracy and completeness, as evidenced by significant improvements across metrics. Meditron-7B shows a remarkable 70.0\% improvement in BERT-Score and a 38.8\% increase in Meteor scores, indicating \textsc{MedAide}'s ability to enhance specialized medical knowledge in smaller models.
(ii) Among frontier models, DeepSeek-R1 with \textsc{MedAide} achieves the highest performance with Meteor scores of 46.54\% and BERT-Score of 17.05\%, outperforming both Claude 3.7 Sonnet and GPT-4o. This suggests that DeepSeek-R1's superior capability in medical knowledge integration is enhanced when augmented with our structured reasoning framework.
(iii) We observe slight performance drops in ZhongJing on ROUGE-L and GLEU, likely due to its limited capacity to handle the RIE phase needed to integrate multifaceted knowledge from retrieval-augmented generation, resulting in sub-optimal prevention guidelines and second opinions.

\noindent \textbf{Results on Medicament Benchmark.}
Table~\ref{medicament} provides the results of the study of different models on the intentions of the composition of the drugs.
(i) Attributed to the drug knowledge injection and prototype-guided embedding in \textsc{MedAide}, the reliability of medication counselling responses is consistently improved across all baselines. DeepSeek-R1 demonstrates significant performance improvements when augmented with \textsc{MedAide}, with its Meteor score increasing from 31.63\% to 67.83\% and its BERT-Score rising from 11.27\% to 43.15\%, showcasing its enhanced ability to integrate structured medication knowledge with recursive reasoning. (ii) Similarly, Baichuan4 achieves a substantial boost, with its GLEU score improving from 7.08\% to 35.99\%, while Claude 3.7 Sonnet shows strong progress in BERT-Score, increasing from 10.52\% to 41.86\%, highlighting \textsc{MedAide}'s effectiveness in improving drug information comprehension and specialized dosage recommendation capabilities.
(iii) In the zero-shot reasoning pattern, Meditron-7B's Meteor and BERT-Score remarkably enhance from 12.29\% to 34.84\% and 2.12\% to 29.43\% respectively, verifying that our decision-analysis module provides favourable factual evidence in drug contraindication understanding, even for smaller medical-specialized models.

\noindent \textbf{Results on Post-Diagnosis Benchmark.}
(i) In the post-diagnosis applications from Table~\ref{PostDiagnosis}, we observe that most \textsc{MedAide}-based models are superior to vanilla baselines by large margins. DeepSeek-R1 demonstrates exceptional improvements with a 28.8\% increase in Meteor score, highlighting its superior capability in generating detailed post-treatment care instructions when augmented with our framework.
(ii) Claude 3.7 Sonnet performs impressively in generating personalized rehabilitation recommendations, with a notable 7.8\% improvement in BERT-Score and a 30.7\% increase in Meteor score, suggesting enhanced capabilities in producing contextually appropriate recovery guidance tailored to individual patient conditions.
(iii) Meanwhile, Baichuan4 shows substantial improvements in the ROUGE-L metric by 43.7\%, revealing the advantages of the \textsc{MedAide}-based version in processing dynamic medical data and producing comprehensive progress tracking reports with better continuity and coherence.

\subsection{Automatic GPT-4o Evaluation}
This study employs GPT-4o to evaluate how \textsc{MedAide} enhances LLMs across reasoning and non-reasoning architectures. The framework differentially augments core capabilities: for reasoning models like DeepSeek-R1, it strengthens clinical reasoning pathways, particularly in complex diagnostic interpretation and longitudinal case analysis. For non-reasoning models, it improves structured knowledge retrieval in domains like medication management. Most significantly, \textsc{MedAide} appears to create synergistic effects, not merely boosting performance metrics but changing how models process medical information, as evidenced by its distinct enhancement patterns across architectures. The persistent challenges in pharmacological tasks suggest these require specialized knowledge integration beyond general reasoning or retrieval improvements, pointing to a critical direction for future medical AI development.

\begin{table}[t]
\centering
\renewcommand{\arraystretch}{1.2}
\setlength{\tabcolsep}{40pt}
\caption{Comparison of F1 scores (\%) across five benchmarks. ``GPT-4o R'' means the GPT-4o Recognition.}
\resizebox{1.0\linewidth}{!}{%
\begin{tabular}{c|c|c|c}
\toprule
{\textbf{Benchmarks}}        & {\textsc{MedAide}}      & { w/o RIE} & { w/ GPT-4o R} \\ \midrule
{ Pre-Diagnosis}  & { \textbf{0.76}} & { 0.62}                & { 0.47}                \\ 
{ Diagnosis}      & { \textbf{0.83}} & { 0.61}                & { 0.49}                 \\ 
{ Medicament}     & { \textbf{0.80}} & { 0.49}                & { 0.51}                 \\ 
{ Post-Diagnosis} & { \textbf{0.63}} & { 0.56}                & { 0.37}                \\ 
Intent Aggregation                    & { \textbf{0.86}} & { 0.60}                & { 0.60}                 \\ \bottomrule
\end{tabular}
}
\label{F1_Analysis}
\vspace{-8pt}
\end{table}

\subsection{Expert Doctor Evaluation}
Expert evaluations play a key role in the practical applications of medical models. We invite 6 doctors (each paid \$300) to select the winners of responses generated by different models before and after the introduction of \textsc{MedAide} by majority voting rule. The response content is holistically evaluated by considering factual accuracy, recommendation practicality, and humanistic care.
(i) As Figure~\ref{vis} shows, all the \textsc{MedAide}-based models exhibit more win rates in different benchmark tests, indicating the effectiveness and applicability of the proposed framework. (ii) \textsc{MedAide} not only improves the healthcare specialization of medical LLMs with different sizes, but also enhances the coping ability of general-purpose models when dealing with complicated medical tasks.

\subsection{Ablation Study}
In this section, we conduct systematic ablation studies to investigate the effects of various components, as shown in Table~\ref{Ablation}.

\noindent \textbf{Necessity of RIE.}
Firstly, we remove the RIE to assess its impact on performance. (i) The observed significant declines in all metrics indicate that RIE is crucial for ensuring that the input information is accurately interpreted by downstream modules with respect to intents. (ii) Incorporating factual information as context enhances the model's ability to accurately comprehend user queries, mitigating hallucinations and reducing ambiguous interpretations.

\noindent \textbf{Importance of IPM.}
We replace the learning-based contextual encoder with the prompt-based GPT-4o to explore the impact of different intent recognition strategies. (i) Despite the improvement over the vanilla baseline, GPT-4o recognition is a suboptimal solution. We argue that the encoder trained with explicitly supervised information can align more purposefully to actual medical intents, producing more personalized and representative judgments than prompt engineering.
(ii) Our default strategy has better flexibility, which can be dynamically optimized according to specific healthcare scenarios.

\noindent \textbf{Effectiveness of RAC.}
Furthermore, we directly assemble the different outputs produced by the activated agents to serve as an alternative candidate. The results from the bottom of Table~\ref{Ablation} show consistent performance drops of the model on different medical intent understandings, proving the effectiveness of our method.
A plausible explanation is that the decision analysis module not only summarizes the different agent outputs in an organized manner, but also provides comprehensive and accurate conclusions based on medical guidelines and patient histories.

\begin{table}[t]
\renewcommand{\arraystretch}{1.2}
\setlength{\tabcolsep}{30pt}
\caption{Comparison of BLEU-1/ROUGE-L scores (\%) between \textsc{MedAide} and other collaboration frameworks.}
\centering
\resizebox{1\linewidth}{!}{%
\begin{tabular}{c|c|c|c}
\toprule
{\textbf{Benchmarks}}        & {\textsc{MedAide}}              & {MedAgents}    & {MDAgents-Group}             \\ \midrule
{ Pre-Diagnosis}  & { \textbf{15.93/17.64}} & { 14.33/15.26}  & { 15.74/17.21}          \\ 
{ Diagnosis}      & { \textbf{27.80/23.67}} & { 25.46/19.54}    & { 27.04/23.45}       \\ 
{ Medicament}     & { \textbf{45.13/52.17}} & { 28.69/36.07}     & { 43.37/49.58}      \\ 
{ Post-Diagnosis} & { \textbf{26.83/18.12}}          & { 25.45/16.69}  & { 25.94/18.11} \\ \bottomrule
\end{tabular}
}
\label{Comparision_Analysis}
\vspace{-8pt}
\end{table}

\subsection{Intent Detection Analysis}
To further observe the performance of intention detection, Table~\ref{F1_Analysis} compares the F1 scores computed between the intents recognized by \textsc{MedAide} and ground truths under different strategies. We randomly sample 100 instances from each benchmark to compose a more challenging intention aggregation benchmark. The core observations are as follows. 
(i) The complete framework achieves the best results, reaching even an 86\% performance on the aggregate benchmark, exhibiting omni-intention semantic understanding capabilities. 
(ii) Significant performance deteriorations occur at ``w/o QR'' and ``w/ GPT-4o R'', implying the effectiveness of fine-grained decomposition of medical queries and the development of our customized intent recognition mechanism.

\begin{figure*}[t]
  \centering
  \includegraphics[width=\linewidth]{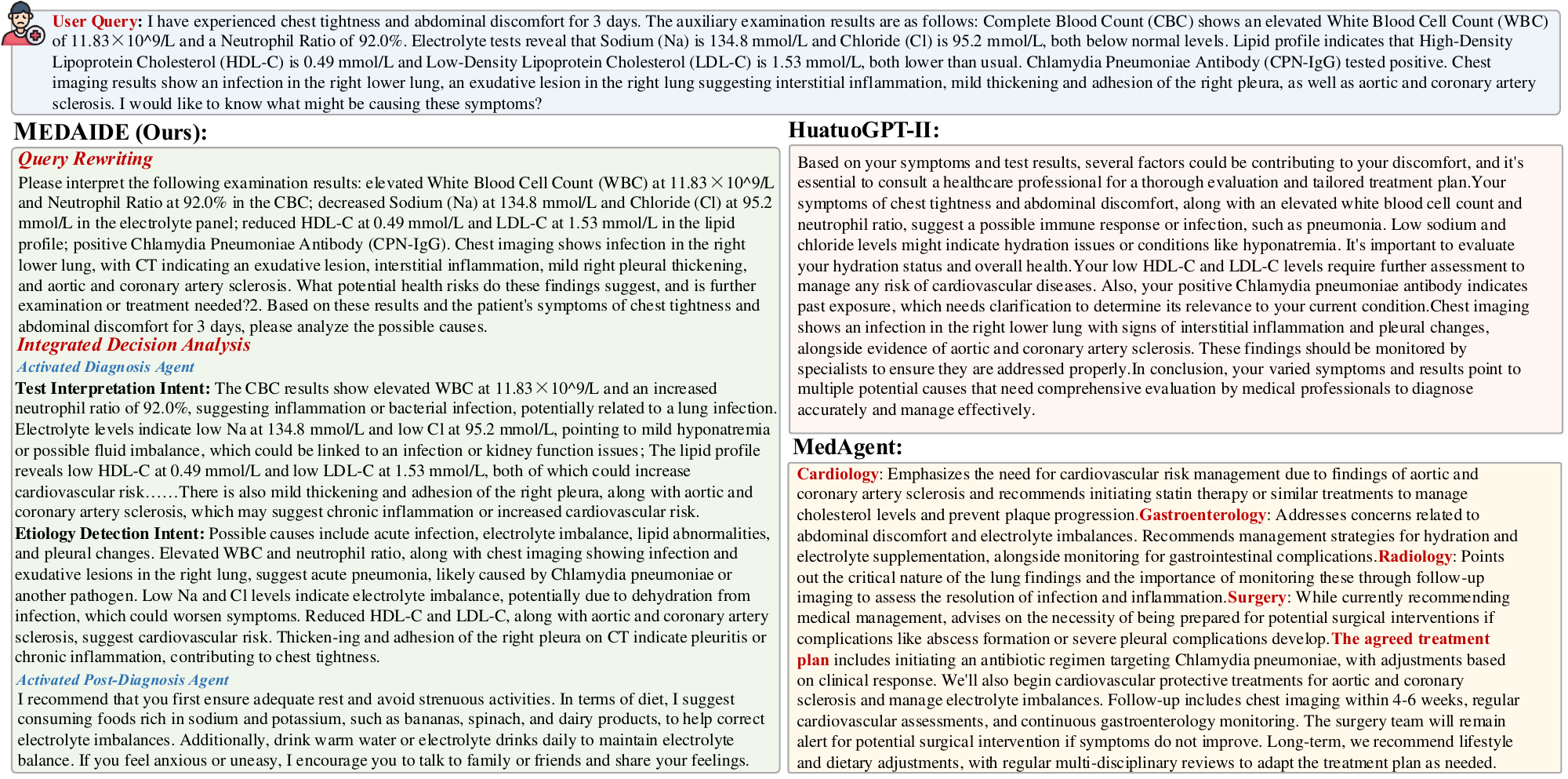}
  \caption{Response visualization analysis on different models for the same user query.
  }
  \label{case_study}
\end{figure*}

\begin{figure}[t]
  \centering
  \includegraphics[width=0.5\linewidth]{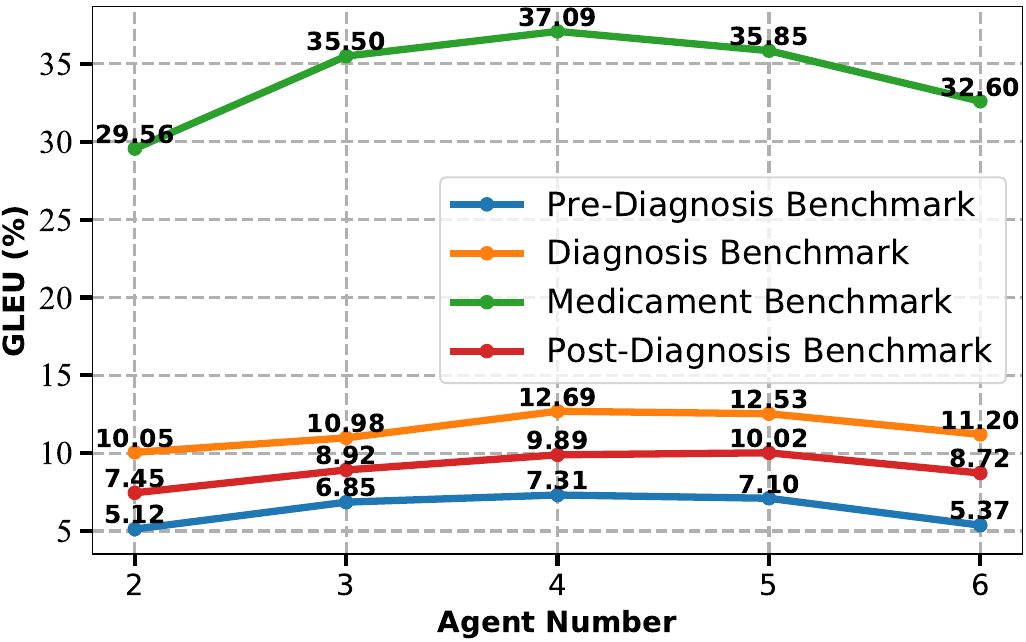}
  \caption{The performance of different agent groups, formed by varying numbers of agents, across the four medical benchmarks.
  }
  \label{agent_num}
\end{figure}

\subsection{Collaboration Framework Comparison}
Here, we compare the reproducible multi-agent framework MedAgents~\cite{tang2024medagentslargelanguagemodels} and MDAgents~\cite{kim2024mdagentsadaptivecollaborationllms}. MedAgents dynamically coordinates multidisciplinary medical experts across various specialties, with each expert system implemented based on GPT-4o architecture. These domain specialists collaboratively analyze cases to produce unified diagnostic reports. For optimal performance-efficiency balance, the system employs 6 active domain experts per case, while the pre-configured Group mode ensures top-tier clinical decision-making through optimized specialist combinations.
(i) From Table~\ref{Comparision_Analysis}, our framework outperforms MedAgents and MDAgents on all four benchmarks. For example, on the Medicament benchmark, \textsc{MedAide} beats MedAgents by large margins with absolute gains of 16.44/16.70\% on the BLEU-1/ROUGE-L scores.
This strength comes from our specialized drug retrieval, which provides a refined context of drug information for the model.
(ii) Also, the benefits of Pre-Diagnosis and Post-Diagnosis tasks reflect that the proposed \textsc{MedAide} captures the key characteristics of diagnostic and nursing demands more efficiently.

\subsection{Response Visualization Analysis}
To intuitively compare the response quality of the different models, in Figure~\ref{case_study} we show the results generated by the \textsc{MedAide} and SOTA methods under the same user query.
HuatuoGPT-II provides extensive medical knowledge, but lacks targeted recommendations. Although the MedAgents framework gives detailed treatment solutions through multidisciplinary perspectives, it lacks coherent diagnostic reasoning.
In contrast, our framework interprets patient outcomes in detail and provides personalized diagnoses and subsequent treatments. \textsc{MedAide} shows favorable usefulness in the management of electrolyte imbalance and cardiovascular risk.

\subsection{Effect of Task Decomposition Granularity}
To systematically analyze how task decomposition granularity affects performance, we design a controlled experiment where the original 17 healthcare intents are dynamically allocated to agents in different stage configurations (ranging from 2 to 6 stages). Each agent is assigned to handle a specific stage, with intent coverage adjusted according to stage divisions (\textit{e.g.,} 2-stage: Pre-Diagnosis and Post-Diagnosis; 4-stage: Pre-Diagnosis, Diagnosis, Medication, and Post-Diagnosis).
As shown in Figure~\ref{agent_num}, \textsc{MedAide} achieves optimal performance with 4 decompositions, where intent-specific responsibilities are balanced without overlap. Progressive improvements were observed when increasing from 2 to 4 agents, confirming that finer task decomposition enhances multi-intent handling. However, further subdivision (5–6 stages) yields diminishing returns, as excessive granularity introduces coordination overhead.

\section{Conclusion}
This paper introduces \textsc{MedAide}, a multi-agent framework for complex medical scenarios. Leveraging Regularization-guided infomation extraction module, Dynamic intent prototype matching module, and Rotation agent collaboration mechanism, \textsc{MedAide} enhances the model's comprehension of medical intents and demonstrates effectiveness across sophisticated medical benchmarks.

\noindent \textbf{Limitations.}
The \textsc{MedAide} framework, while making significant strides in integrating large-scale medical agents with real clinical environments, has its limitations. Currently, the framework incorporates 26,684 drug samples and 506 genuine clinical case records. Despite this scope, it lacks comprehensive coverage, particularly in research on rare diseases and specialized medications. Furthermore, as the initial versions primarily emphasized linguistic processing, future research aims to extend into multimodal capabilities, particularly the integration of medical imaging data, to enhance the framework's ability to process visual information in clinical diagnostics. The reliance on OpenAI's API may pose potential operational challenges, suggesting that future studies should explore using more efficient open-source models as viable alternatives.

\noindent \textbf{Ethics Consideration.}
We are acutely aware of the necessity for privacy and data protection. All data utilized has undergone thorough de-identification, with all sensitive information removed, and verified by a partnering medical institution.
We invite doctors to perform only evaluations of model responses without involving any form of human subject research. All participants are compensated \$300 for their work, which strictly adheres to the minimum hourly rate for the region in which the work is performed.
We strictly follow the license agreements of publicly available databases when utilizing healthcare-related data. For the constructed data, we have undergone an internal ethical review by the ethics review board of our partnering medical institutions and are licensed and approved.

\section*{Acknowledgments}
This work is supported in part by the National Key R\&D Program of China (2021ZD0113503) and in part by the Shanghai Municipal Science and Technology Major Project (2021SHZDZX0103).

\clearpage

\bibliographystyle{plainnat}
\bibliography{main}



\end{document}